\newif\ifPrePrint
\PrePrinttrue

\documentclass[letterpaper, 10 pt, conference]{ieeeconf}
\IEEEoverridecommandlockouts                              

\def\baselinestretch{0.97}

\usepackage[switch]{lineno}
\usepackage[noadjust]{cite}
\usepackage{graphics} 
\usepackage{xcolor}
\usepackage{epsfig} 
\usepackage{epstopdf}
\usepackage{amsmath} 
\usepackage{amssymb}  
\usepackage{stmaryrd}
\usepackage{transparent}
\usepackage{import}

\usepackage{siunitx}
\usepackage[font=footnotesize,labelfont=bf]{caption}
\usepackage{subfig}
\usepackage{url}
\usepackage{hyperref}

\usepackage{tikz}
\usepackage{siunitx}

\usepackage{amsthm}
\newtheorem{obj}{Objective}
\newtheorem*{obj*}{Objective}
\newtheorem{prob}{Problem}
\newtheorem*{prob*}{Problem}

\newtheorem{assumption}{Assumption}
\newtheorem*{assumption*}{Assumption}

\usetikzlibrary{shapes,arrows,positioning,calc}
\newcommand{\todo}[1]{\textcolor{red}{TODO: #1}}

\newcommand{\toWeixuan}[1]{\textcolor{red}{Weixuan: #1}}

\newcommand{\norm}[1]{\left\lVert#1\right\rVert}
\newcommand{\mvec}[1]{\boldsymbol{#1}}

\newcommand{\Bnom}{B_\mathrm{nom}}
\newcommand{\Bdist}{B_\mathrm{dist}}
\newcommand{\Ddist}{D_\mathrm{dist}}

\newcommand{\task}{_t}
\newcommand{\refTask}{_t^{r}}

\newcommand{\informative}{_i}
\newcommand{\refInformative}{_i^{r}}

\newcommand{\sampleIndex}{^j}

\newcommand{\sampledStateInputPair}{\mvec{z}\sampleIndex}
\newcommand{\stateRef}{\mvec{x}^r}
\newcommand{\inputRef}{\mvec{u}^r}

\newcommand{\stateTask}{\mvec{x}{\refTask}}
\newcommand{\inputTask}{\mvec{u}{\refTask}}
\newcommand{\stateTaskTraj}{X{\refTask}}
\newcommand{\inputTaskTraj}{U{\refTask}}
\newcommand{\stateInputTaskTraj}{Z{\refTask}}
\newcommand{\inputTaskTrajPerfect}{\inputTaskTraj^*}

\newcommand{\stateTraj}{X}
\newcommand{\inputTraj}{U}
\newcommand{\stateInputTraj}{Z}

\newcommand{\dState}{\delta \mvec{x}}
\newcommand{\dInput}{\delta \mvec{u}}

\newcommand{\dStateTraj}{\Delta X}
\newcommand{\dInputTraj}{\Delta U}

\newcommand{\posState}{\mvec{p}}
\newcommand{\velState}{\dot{\mvec{p}}}
\newcommand{\attState}{\mvec{\eta}}
\newcommand{\angVelState}{\mvec{\omega}}

\newcommand{\stateInf}{\mvec{x}{\refInformative}}
\newcommand{\candidateStateInf}{\mvec{x}_{i, \mathrm{cand}}^{r}}
\newcommand{\inputInf}{\mvec{u}\informative}

\newcommand{\stateInfTraj}{X{\refInformative}}
\newcommand{\candidateStateInfTraj}{X_{i, \mathrm{cand}}^{r}}
\newcommand{\inputInfTraj}{U{\refInformative}}

\newcommand{\inputInstance}{\mvec{u}}
\newcommand{\stateInstance}{\mvec{x}}
\newcommand{\stateInputPairInstance}{\mvec{z}}

\newcommand{\stateInputPairTask}{\mvec{z}\refTask}

\newcommand{\stateSet}{\mathcal{X}}
\newcommand{\inputSet}{\mathcal{U}}
\newcommand{\stateInputSet}{\mathcal{Z}}

\newcommand{\stateSetSampled}{\mathcal{X}'}
\newcommand{\inputSetSampled}{\mathcal{U}'}

\newcommand{\stateInfActual}{\mvec{x}{\informative}}
\newcommand{\inputInfActual}{\mvec{u}{\informative}}

\newcommand{\stateTaskActual}{\mvec{x}{\task}}
\newcommand{\inputTaskActual}{\mvec{u}{\task}}

\newcommand{\stateInputInformativeActualTraj}{Z{\informative}}

\newcommand{\stateInputData}{Z}
\newcommand{\stateInputInfSimTraj}{\bar{Z}\informative}

\newcommand{\stateInfSim}{\bar{\mvec{x}}\informative}
\newcommand{\inputInfSim}{\bar{\mvec{u}}\informative}

\newcommand{\pdfs}[2]{\mathcal{N}_{#1} \! \left( #2 \right) } 

\newcommand{\stateTaskPerfect}{\mvec{x}^*{\task}}
\newcommand{\inputTaskPerfect}{\mvec{u}^*{\task}}

\newcommand{\stateTaskActualTraj}{\tilde{X}{\task}}
\newcommand{\inputTaskActualTraj}{\tilde{U}{\task}}

\newcommand{\stateTaskSim}{\bar{\mvec{x}}\task}
\newcommand{\inputTaskSim}{\bar{\mvec{u}}\task}
\newcommand{\stateInputTaskSimSample}{\bar{\mvec{z}}\task\sampleIndex}
\newcommand{\stateInputInfSimSample}{\bar{\mvec{z}}\informative\sampleIndex}

\newcommand{\stateTaskSimTraj}{\bar{X}{\task}}
\newcommand{\inputTaskSimTraj}{\bar{U}{\task}}

\newcommand{\stateInputTaskSimTrajSampled}{\bar{Z}\task\sampleIndex}

\newcommand{\sampledLearnedModelInput}{\bar{\zeta}}
\newcommand{\learnedModelInput}{\zeta}
\newcommand{\learnedModelInputSet}{\mathcal{Z}}
\newcommand{\learnedModelInputTaskSet}{\mathcal{Z}{\refTask}}
\newcommand{\learnedModelInputTaskDeltaSet}{\mathcal{Z}_{\Delta{\refTask}}}
\newcommand{\learnedModelInputTaskDeltaSetSampled}{\mathcal{Z}'_{\Delta{\refTask}}}
\newcommand{\learnedModelInputTaskDeltaSetEstimate}{\hat{\mathcal{Z}}_{\Delta{\refTask}}}

\newcommand{\discreteState}[1]{\boldsymbol{x}[#1]}
\newcommand{\discreteInput}[1]{\boldsymbol{u}[#1]}
\newcommand{\feedbackInput}[1]{\boldsymbol{u}_\mathrm{fb}[#1]}

\newcommand{\D}{\mathrm{d}}

\newcommand{\statePerf}{\mvec{x}_{\mathrm{perf}}}
\newcommand{\inputPerf}{\mvec{u}_{\mathrm{perf}}}

\newcommand{\feedbackCtrl}[1]{\pi(#1)}
\newcommand{\feedbackCtrlNom}[1]{\pi_\mathrm{nom}(#1)}

\newcommand{\forceVecCmd}{\mvec{f}_{\mathrm{cmd}}}

\newcommand{\torqueVecCmd}{\mvec{\tau}_{\mathrm{cmd}}}

\newcommand{\R}{\mathbb{R}}
\newcommand{\N}{\mathbb{N}}

\newcommand{\noiseVar}{\sigma^2}
\newcommand{\GPobservation}{z}
\newcommand{\GPobservationStacked}{\mvec{Z}}
\newcommand{\GPinput}{\mvec{\xi}}
\newcommand{\GPinputStacked}{X}
\newcommand{\GPtestInput}{\mathrm{\mvec{\xi}}^*}

\newcommand{\modeledNonlinearDyn}{h}
\newcommand{\unkownNonlinearDyn}{g}
\newcommand{\unkownNonlinearDynElement}[1]{g_{#1}}
\newcommand{\estimateUnkownNonlinearDyn}{\hat{g}}
\newcommand{\trueNonlinearDyn}{f}
\newcommand{\estimateNonlinearDyn}{\hat{f}}

\newcommand{\stateVec}{\mvec{x}}
\newcommand{\stateVecD}{\mvec{x}[k]}
\newcommand{\traj}[2]{(#1 )}
\newcommand{\stateVecDNext}{\mvec{x}[k+1]}

\newcommand{\GPmean}[1]{m(#1)}

\newcommand{\GPpredictionMean}[1]{\mu(#1)}
\newcommand{\sampledState}{s}
\newcommand{\sampledResidualDynamics}{\mvec{g}'}
\newcommand{\priorMean}[1]{\mu(#1)}
\newcommand{\priorMeanElement}[2]{\mu_{#1}(#2)}
\newcommand{\priorVariance}[1]{\sigma^2(#1)}
\newcommand{\variance}[1]{\sigma^2(#1)}
\newcommand{\varianceElement}[2]{\sigma^2_{#1}(#2)}

\newcommand{\posteriorVariance}[2]{\sigma_{#1}^2 (#2)}

\newcommand{\GPpredictionVar}[1]{\sigma(#1)}
\newcommand{\GPpredictionMeanSingle}[2]{\mu_{#1}(#2)}
\newcommand{\GPpredictionVarSingle}[2]{\sigma^2_{#1}(#2)}
\newcommand{\unknownDynamicsFunc}[1]{g(#1)}

\newcommand{\inputVec}{\mvec{u}}
\newcommand{\inputVecD}{\mvec{u}[k]}

\newcommand{\figref}[1]{Fig.~\ref{#1}}

\newcommand{\stateInfParam}{\mvec{\Theta}_{\stateInstance}}
\newcommand{\stateInfParamSet}{\mathcal{O}_{\stateInstance}}

\newcommand{\ve}[1]{\mvec{e}_{#1}}

\newif\ifold
\oldfalse

\newif\ifskeleton
\skeletonfalse

\newif\ifrevision
\revisionfalse
\ifrevision
  \newcommand{\rev}[1]{\textcolor{red}{#1}}
  
\else
  \newcommand{\rev}[1]{#1}
  
\fi

\ifPrePrint
\makeatletter
\def\ps@titlepagestyle{
	\def\@oddfoot{}\def\@evenfoot{}
	\def\@oddhead{\textcolor{red}{\sf\footnotesize Preprint version, final version at http://ieeexplore.ieee.org/ \hfill IEEE International Conference on Robotics and Automation 2021}}
	\def\@evenhead{\textcolor{red}{\sf\footnotesize  Preprint version, final version at http://ieeexplore.ieee.org/  \hfill IEEE International Conference on Robotics and Automation 2021}}%
}%
\makeatother
\makeatletter
\def\ps@headings{
	\def\@oddfoot{\textcolor{red}{\sf\footnotesize  Preprint version, final version at http://ieeexplore.ieee.org/ \qquad\qquad\quad \thepage \;\;~\hfill ~\hfill IEEE International Conference on Robotics and Automation 2021}}\def\@evenfoot{\hfill\thepage\hfill}
	\def\@oddhead{}\def\@evenhead{}%
}%
\makeatother
\fi	

\begin{document}
\pagenumbering{gobble}

%

\title{\LARGE \bf Sampling-based trajectory generation for active model learning of dynamical systems}
\author{Weixuan Zhang, Lionel Ott, Marco Tognon, Roland Siegwart, and Juan Nieto
\thanks{This work was supported by the NCCR Digital Fabrication.}
\thanks{Authors are with the Autonomous Systems Lab, ETH Z\"{u}rich, Leonhardstrasse 21, 8092 Zurich, Switzerland. e-mail:
        {\tt wzhang@mavt.ethz.ch}}
}

\title{\LARGE \bf Active Model Learning using Informative Trajectories for Improved\\ Closed-Loop Control on Real Robots}
\author{Weixuan Zhang,  Marco Tognon, Lionel Ott, Roland Siegwart, and Juan Nieto
\thanks{This work was supported by the NCCR Robotics, NCCR Digital Fabrication and Armasuisse.}
\thanks{Authors are with the Autonomous Systems Lab, ETH Z\"{u}rich, Leonhardstrasse 21, 8092 Zurich, Switzerland. e-mail:
        {\tt wzhang@mavt.ethz.ch}}
}

\maketitle

\ifPrePrint
\else
	\thispagestyle{plain}
	\pagestyle{plain}
\fi

\begin{abstract}
Model-based controllers on real robots require accurate knowledge of the system dynamics to perform optimally.
For complex dynamics, first-principles modeling is not sufficiently precise, and data-driven approaches can be leveraged to learn a statistical model from real experiments.
However, the efficient and effective data collection for such a data-driven system on real robots is still an open challenge. 
This paper introduces an optimization problem formulation to find an informative trajectory that allows for efficient data collection and model learning. 
We present a sampling-based method that computes an approximation of the trajectory that minimizes the prediction uncertainty of the dynamics model.
This trajectory is then executed, collecting the data to update the learned model.
%
We experimentally demonstrate the capabilities of our proposed framework when applied to a complex omnidirectional flying vehicle with tiltable rotors.
Using our informative trajectories  results in models which outperform models obtained from non-informative trajectory by 13.3\% with the same amount of training data. 
Furthermore, we show that the model learned from informative trajectories generalizes better than the one learned from non-informative trajectories, achieving better tracking performance on different tasks.

\end{abstract}

\setcounter{section}{0}
\section{Introduction}

Model-based controllers have shown to be useful in various robotics applications. 
Especially when accurate models are available, these controllers can exhibit impressive performance \cite{abbeel2010autonomous}, \cite{kamel2017model}. \rev{Compared to model-free methods such as reinforcement learning, there is no need of training samples to train a control policy.}
On the other hand, it can be hard to obtain a good dynamical model for complex systems such as humanoid robots \cite{kuindersma2016optimization}, race cars on uneven terrains \cite{ostafew2014learning}, soft robots \cite{gillespie2018learning}, and novel fully actuated multi-rotor flying vehicles~\cite{bodie2019omnidirectional} like the one considered in this work (see \figref{pic:omav}).

One approach to solve the modeling problem is to rely on learning techniques: Through interaction with the real-world and data collection a statistical dynamics model is trained, which is either directly fed into a model-based controller, e.g. \cite{kabzan2019learning,ostafew2014learning,zhang2020learning,nguyen2010using}, or used in simulation to train a control policy \cite{hwangbo2019learning}.
One challenge for these approaches is that often the training data has a different distribution than the test data due to several reasons: first, model uncertainties and feedback controller might lead the system to a state not encountered in a previous data collection routine. 
Secondly, given partial model knowledge, the region of the data that leads to the best performance is a-priori unknown.
Finally, the closed-loop dynamics change as the model used by the controller is updated.  
One could perform a large number of experiments to cover as much of the input space as possible during training. However, for robotic systems with high-dimensional and continuous state space, the search space typically is too large to be searched exhaustively.
Furthermore, the dynamics can change significantly during consecutive experiments, e.g., the crash of a flying vehicle could damage its motors and invalidate the previous training data.
Even when considering a specific task, a good model is required in the working area of the state and input spaces\rev{, which still might be large.} Thus, it is desirable to have an efficient scheme to collect training data locally around the desired task \rev{if a precise enough first-principle parametric model is not available or hard to obtain}.
\rev{As these learning techniques are nonparametric, common tools from parametric system identification \cite{ljung1999system}, e.g., persistence of excitation, are not applicable.}

One idea is to use the statistical information learned from training data to infer the region where to sample data, thus improving sampling efficiency. This is a well-known approach in machine learning called \textit{active learning} \cite{settles2009active}.
In this paper we exploit such an idea: we rely on the previously learned statistical model to get an estimate of the region of interest. We then generate an informative trajectory that reduces the overall uncertainty in the estimated region. This trajectory is then executed in the real world to collect data.

More specifically, in a first step, possible informative locations are inferred in simulation from the previously learned model. 
Then, different informative trajectories are sampled and evaluated according to a cost metric, which is defined as the integral of the predictive uncertainty over these possible locations.
The most informative trajectory is then selected and executed on the real robot to collect the data.
As a result, the model learned from this informative trajectory should result in improved control performance and a better generalization.
The latter is achieved because the informative trajectory reduces the uncertainty over a large region of state and input space.


The contributions of this paper are summarized as follows:
\begin{itemize}
  \item A formal mathematical formulation of the problem of efficient data collection for learning dynamics model.
  \item A practical strategy to \rev{\textit{efficiently}} collect task-relevant data that improves the \rev{model-based} control performance when used to update the learned model. 
  \item Real experimental results conducted on a complex overactuated omnidirectional flying system with nonlinear dynamics and 18 actuators.
  \rev{For a figure-8 trajectory, two runs of trajectory flight lead to an angular acceleration tracking error reduction of 54.4\% }
\end{itemize}

\subsection{Related work}

Active learning  in robotics is mostly defined in a regression setting: a regression mapping between an input and an output space is to be learned while the sample complexity is minimized. The exploration of the sample space is typically driven by some metric often consisting in variants of the expected informational gain. 

Considering active dynamics model learning, existing work includes the use of information gain on parameter estimates (\cite{schrangl2020iterative,wilson2016dynamic}), Gaussian processes (\cite{koller2018learning,buisson2019actively, zimmer2018safe}), and neural networks \cite{nakka2020chance}.
They typically generate trajectories that minimize a defined metric, trading off between exploration and exploitation.
 Aside from the parameter estimates approach, little  work is done on real robots.
%
%
%

We can also distinguish approaches depending whether the trajectory generation is performed online or offline.   The online approaches are often done in a receding horizon fashion \cite{borrelli2017predictive}, where trajectories are regenerated at a certain frequency  on the fly during experiments. 
This constant update helps reducing the distance between the desired inputs and achieved ones. 
However, this approach is  computationally intensive. 
While exploring a state of interest, the robot cannot always stay stationary waiting for a new planned trajectory. 
Up to date, this method exists only in theoretical works validated in simulation~\cite{koller2018learning, buisson2019actively, capone2020localized}.


The offline approach has the shortcoming that the planned trajectory has a larger distance to the executed one, but applicable on real robots. 
In \cite{nakka2020chance}, the trajectory generation is formulated as a variable-constrained problem and validated on a simulated overactuated robotic spacecraft.
In \cite{zimmer2018safe}, the input trajectories are parametrized by consecutive trajectory sections and the most informative and safe trajectory is then executed. Their formulation did not take into account closed-loop control. The method is applied on a high-pressure fluid injection system. 
Our investigation belongs to this approach: we make use of the previously learned model and simulations to reduce the deviation of the executed trajectory to the desired one.
In this work, we demonstrate that this approach works for complex robots and efficiently improve control performance.

\section{Modeling and Problem Statement} \label{sec:problemStatement}
We consider a generic system whose dynamics in the discrete time domain are described by:
\begin{equation}
  \discreteState{k+1} = \trueNonlinearDyn(\discreteState{k}, \discreteInput{k}),
  \label{eqn:dynamics:true}
\end{equation}
where $\trueNonlinearDyn(\cdot, \cdot)$ is a Lipschitz-continuous function\footnote{This is a common assumption that does not limit the validity of the work since most of the considered robotic systems have Lipschitz-continuous dynamics.} and represents the \textit{true dynamics}. 
$\discreteState{k} \in \stateSet \subset \R^n$ and $\discreteInput{k} \in \inputSet \subset \R^m$ describes the state and the control input of the dynamical system at time $k \in \N_{\geq 0}$. 
To simplify the notation, $\discreteState{k}$ denotes $\stateVec(kT)$ where $T \in \R_{> 0}$ is the sampling time.
We remark that \rev{a perfect knowledge of} $\trueNonlinearDyn(\cdot, \cdot)$ is in general not available. 
We might have only an estimation of it denoted by $\estimateNonlinearDyn(\cdot, \cdot)$.

The considered task consists in a trajectory tracking problem. 
A desired \textit{task state trajectory} is defined by the sequence of state values $\stateTaskTraj = (\stateTask[0], \,  \ldots , \stateTask[N])$ in the time horizon $N \in \N_{> 0}$. \rev{Throughout this paper, we use a capitalized letter to indicate a sequence of vectors with a time horizon of $N$.
The subscript $\star_t$ is used to denote the quantities related to the task trajectory tracking problem, while the superscript $\star^r$ denotes reference state or input.
We first introduce the following assumption
\begin{assumption}\label{assumption:perfectTracking}
A model-based controller $\feedbackCtrl{\cdot, \cdot, \cdot}$ that is a function of a reference state $\stateRef[k]$, a state $\stateInstance[k]$, and an estimated dynamics model $\estimateNonlinearDyn$ is provided
\begin{equation}\label{eq:feedback_ctrl}
  \inputInstance[k] = \feedbackCtrl{\stateRef[k], \stateInstance[k], \estimateNonlinearDyn}. 
\end{equation}
Furthermore, if $\discreteState{0} = \stateTask[0]$, and $\estimateNonlinearDyn(\stateInstance, \inputInstance) = \trueNonlinearDyn(\stateInstance, \inputInstance)$ for every $(\stateInstance, \inputInstance) \in \stateInputSet = \stateSet \times \inputSet$,
then
\begin{equation}
\stateTaskActual[k+1] = \trueNonlinearDyn(\stateTaskActual[k], \feedbackCtrl{\stateTask[k], \stateTaskActual[k], \estimateNonlinearDyn}) = \stateTask[k+1],
\end{equation} 
for every $k = 0, \ldots, N-1$. 
\end{assumption} 
}
This condition describes a perfect tracking of the desired trajectory 
\rev{given a perfect modeling.}
\rev{We further remark that the given task state trajectory is feasible, so that there exists at least one task input trajectory to achieve it}
We define the sequence of inputs that provides perfect tracking as $\inputTaskTraj = (\inputTask[0], \ldots, \inputTask[N])$, called \textit{task input trajectory}.
%

\begin{obj}\label{obj}
	Considering the closed-loop system \eqref{eqn:dynamics:true} and \eqref{eq:feedback_ctrl}, our objective is to define an active learning method aiming at optimizing the data collection process to 
\begin{itemize}
	\item make it more efficient (less experiments and data points),
	\item improve the precision of the learned model,
	\item improve the generalizability of the learned model,
	\item minimize the tracking error.  
\end{itemize}
\end{obj}
We shall show how the learning problem can be reformulated to address such objectives.

Without loss of generality, we can decompose the true dynamics into two components:
\begin{align}
	\trueNonlinearDyn(\discreteState{k}, \discreteInput{k}) = \modeledNonlinearDyn(\discreteState{k}, \discreteInput{k}) + \unkownNonlinearDyn(\discreteState{k}, \discreteInput{k}),
	\label{eq:dynamcis:true:decomposition}
\end{align}
where $\modeledNonlinearDyn(\cdot, \cdot)$ is called \textit{first principles dynamics}, corresponding to the model reflecting physical laws. We consider $\modeledNonlinearDyn(\cdot, \cdot)$ to be known.
  $\unkownNonlinearDyn(\cdot, \cdot)$ is  called \textit{residual dynamics}, corresponds to all other elements not modeled by $\modeledNonlinearDyn$. $\unkownNonlinearDyn$ is assumed unknown and we only have an estimation denoted by $\estimateUnkownNonlinearDyn$.
    
This modeling allows to exploit the knowledge we already have about the system, reducing the learning effort and making it possible to employ several model-based controllers. 

Once again, it is clear that, considering the control law \eqref{eq:feedback_ctrl} with $\estimateNonlinearDyn = \modeledNonlinearDyn + \estimateUnkownNonlinearDyn$, the closed loop system achieves perfect tracking if $\estimateUnkownNonlinearDyn(\stateInputPairInstance) = \unkownNonlinearDyn(\stateInputPairInstance)$ for every $\stateInputPairInstance := (\stateInstance, \inputInstance) \in \stateInputSet$.
\rev{For simplicity we use $\stateInputPairInstance$ to denote the state-input pair $(\stateInstance, \inputInstance)$.}
\rev{We assume that a Bayesian prior model \cite{congdon2007bayesian} over the residual dynamics is given.} 
That is, for a given test point $\stateInputPairInstance$, the belief of the value of $\unkownNonlinearDyn(\stateInputPairInstance)$ follows a Gaussian probability distribution $\pdfs{\stateInputPairInstance}{\priorMean{\stateInputPairInstance}, \priorVariance{\stateInputPairInstance}}$.
We denote the mean and variance of $\pdfs{\stateInputPairInstance}{\cdot, \cdot}$ as $\priorMean{\stateInputPairInstance} \in \R{}^n$ and $\priorVariance{\stateInputPairInstance} \in \R{}_{\geq 0}^{n \times n}$, respectively. 
Note that the distribution is a function of the test point $\stateInputPairInstance$.
We consider the estimation of the residual dynamics as $\estimateUnkownNonlinearDyn(\stateInputPairInstance) = \priorMean{\stateInputPairInstance}$, which brings to $\estimateNonlinearDyn(\stateInputPairInstance) = \modeledNonlinearDyn(\stateInputPairInstance) + \priorMean{\stateInputPairInstance}$.

\rev{
Let $\stateInputTraj$ denote the state and input trajectory pair $\traj{\stateTraj, \inputTraj}{N}$. The prior model can be updated to a posterior model from trajectory data subsampled from $\stateInputTraj$. In particular, the updated model is described by the posterior mean $\priorMean{\stateInputPairInstance | \stateInputTraj}$ and posterior variance $\variance{\stateInputPairInstance | \stateInputTraj}$.}

\rev{ 
We then introduce the following assumption for the Bayesian model:
\begin{assumption} \label{assumption:uncertainty}
Given two sets of data from trajectory $\stateInputData_1$ and $\stateInputData_2$, for all $\stateInputPairInstance \in \stateInputSet$ and $j = 0, \dots,  n$, $\varianceElement{j}{\stateInputPairInstance | \stateInputData_1} < \varianceElement{j}{\stateInputPairInstance | \stateInputData_2}$ leads to $|\priorMeanElement{j}{\stateInputPairInstance | \stateInputData_1} - \unkownNonlinearDynElement{j}(\stateInputPairInstance)| < |\priorMeanElement{j}{\stateInputPairInstance | \stateInputData_2} - \unkownNonlinearDynElement{j}(\stateInputPairInstance)|$. Furthermore, if $\varianceElement{j}{\stateInputPairInstance | \stateInputData_1}$ approaches zero, $|\priorMeanElement{j}{\stateInputPairInstance | \stateInputData_1} - \unkownNonlinearDynElement{j}(\stateInputPairInstance)|$ approaches zero. The subscript $*_j$ is used to denote the $j$-th vector element or $j$-th diagonal element.
\end{assumption}
The intuition behind this practical assumption is that a high-quality observation (which is possible in robotic applications) near the test point reduces the uncertainty at the test point and therefore reduces the estimation error.}


To improve the knowledge of $\estimateUnkownNonlinearDyn$, suitable data must be collected.
A possible solution is to simply run the task trajectory, over and over, until sufficient data is collected to obtain a good model around $\traj{\stateTaskTraj, \inputTaskTraj}{N}$, or $\stateInputTaskTraj$. 
However, this would require many trials to ensure the collected data is informative enough. 

Departing from this basic approach, here we aim to design an algorithm that automatically derives reference state trajectories $\stateInfTraj$, called \textit{informative state trajectories}. 
\rev{These trajectories aim to efficiently collect data to improve the prior model,} thus reducing the task trajectory tracking error when using the control law \eqref{eq:feedback_ctrl}. 
\rev{
Let $\stateInfActual[k]$ and $\inputInfActual[k]$ denote the inputs and states obtained letting the closed-loop system evolve using $\stateInfTraj$ as reference trajectory. In details
\begin{align}
  \stateInfActual[k+1] &= \trueNonlinearDyn(\stateInfActual[k], \inputInfActual[k])  \nonumber\\ 
  \inputInfActual[k] &= \feedbackCtrl{\stateInf[k], \stateInfActual[k], \estimateNonlinearDyn},\label{eq:realDynamics1}
\end{align}
with $\stateInfActual[0] = \stateInf[0]$.
The subscript $*_i$ is used to denote quantities related to the informative trajectory tracking problem.
}

The following problem is then formulated:
\begin{prob} \label{problem:tracking}
Find $\stateInfTraj$ as solution of:
\begin{equation}
	\begin{aligned}
		\min_{\stateInfTraj} \quad & \sum_{k=0}^{N} \norm{ \stateTask[k] - \stateTaskActual[k] }_2^2 	 \\
		\textrm{s.t.} \quad	
			& \stateTaskActual[k+1] = \trueNonlinearDyn(\stateTaskActual[k], \inputTaskActual[k]) \\
			& \inputTaskActual[k] = \feedbackCtrl{\stateTask[k], \stateTaskActual[k], \estimateNonlinearDyn_\mathrm{posterior}} \\
			& \estimateNonlinearDyn_\mathrm{posterior} = \modeledNonlinearDyn + \estimateUnkownNonlinearDyn \\
			& \estimateUnkownNonlinearDyn(\stateInputPairInstance) = \priorMean{\stateInputPairInstance | \stateInputInformativeActualTraj }, \,
      \stateInputInformativeActualTraj \text{ as in } \eqref{eq:realDynamics1}. \\
	\end{aligned}
	\label{eq:optimization:trackingError}
\end{equation}
\end{prob}

\section{Generation of Informative Trajectories}\label{sec:informativeTrajGeneration}
\rev{This section introduces an optimization problem aiming at minimizing an informative cost metric, the solution of which is equivalent to the solution of \eqref{eq:optimization:trackingError}.}
The problem is solved by practical approximations leading to a sampling-based trajectory generation algorithm.

\subsection{Minimization of the informative cost}
Solving \eqref{eq:optimization:trackingError} is definitely not a trivial problem, even using sampling-based methods.
In fact, since we do not know $\trueNonlinearDyn$, solving \eqref{eq:optimization:trackingError} would require to run two experiments for every sampled informative state trajectory  $\stateInfTraj$, using as reference firstly $\stateInfTraj$ and then $\stateTaskTraj$.

In order to make the problem feasible from a practical point of view, let us recall that using the model-based controller \eqref{eq:feedback_ctrl}, we can achieve perfect tracking by having the perfect knowledge of $\estimateUnkownNonlinearDyn{}$ for all $\stateInputPairInstance \in \learnedModelInputTaskSet$ where
\begin{align}
	\begin{split}
	\learnedModelInputTaskSet = & \{\stateInputPairInstance \in \stateInputSet \, | \, \exists \, k \in (0,\ldots,N) \text{ s.t. } \stateInputPairInstance = \stateInputPairTask[k]\},
	\end{split}
\end{align}
contains the pairs state/input that achieve perfect tracking of the task trajectory.

\rev{According to Assumption \ref{assumption:uncertainty}, } a possible idea is to improve the model by minimizing the uncertainty of the prior model, i.e., $\priorVariance{\stateInputPairInstance}$ for all $\stateInputPairInstance \in \learnedModelInputTaskSet$. 
Thus, we reformulate \eqref{eq:optimization:trackingError} as
\begin{equation}
		\min_{\stateInfTraj} \sum_{ \learnedModelInputTaskSet } \priorVariance{\stateInputPairInstance | \stateInputInformativeActualTraj}.	
	\label{eq:optimization:variance:trackingTraj}
\end{equation}
Recall that $\stateInputInformativeActualTraj$ are computed as in \eqref{eq:realDynamics1}.
\rev{Note that the solution of \eqref{eq:optimization:variance:trackingTraj} allows to minimize the modeling error (Assumption~\ref{assumption:uncertainty}) which in turns leads to the minimization of tracking error (Assumption~\ref{assumption:perfectTracking}). 
Therefore, the solution of \eqref{eq:optimization:variance:trackingTraj} is also the solution of Problem~\ref{problem:tracking}. }

Notice that we focus on reducing the informative cost on the space relevant to the task instead of the entire state/input space $\stateInputSet$.
%
However, from experimental considerations, we remark that improving the model only in $\learnedModelInputTaskSet$ is not enough to achieve good tracking performance. 
In fact, initial errors, noisy measurements, and external disturbances might make the system deviate from $\stateInputTaskTraj$, visiting pairs input/state not included in $\learnedModelInputTaskSet$ for which the model could be imprecise. 
Therefore, to achieve good tracking also in these non-ideal and more realistic conditions, we propose to improve the learning of the model by solving \eqref{eq:optimization:variance:trackingTraj} not only for the points in $\learnedModelInputTaskSet$, but also for the ones that are sufficiently close, i.e., for all $\stateInputPairInstance \in \learnedModelInputTaskDeltaSet$ where 
\begin{align}
	\begin{split}
	\learnedModelInputTaskDeltaSet = & \{\stateInputPairInstance \in \stateInputSet  \, | \, \exists \, \stateInputPairTask \in \learnedModelInputTaskSet \text{ s.t. } \norm{\stateInputPairInstance - \stateInputPairTask} \leq \epsilon \},
	\end{split}
  \label{eq:optimization:BallDef}
\end{align}
with\footnote{With an abuse of notation, we consider $\norm{\stateInputPairInstance - \stateInputPairInstance_\star} = \norm{[\stateInstance^\top \,\; \inputInstance^\top]^\top - [\stateInstance_\star^\top \,\; \inputInstance_\star^\top]^\top}$. A weighted norm can also be used to normalize the components of state and input vectors.} $\epsilon \in \R{}_{\geq 0}$ being a heuristic that can be tuned to control the exploratory behavior of the informative trajectory.
Problem \eqref{eq:optimization:variance:trackingTraj} becomes:
\begin{equation}
		\min_{\stateInfTraj} \int_{ \learnedModelInputTaskDeltaSet } \priorVariance{\stateInputPairInstance | \stateInputInformativeActualTraj}	d \stateInputPairInstance 
	\label{eq:optimization:variance:ballTrackingTraj}
\end{equation}
From now on, we refer to the objective function to be minimized as \textit{informative cost}. 


%

%

The problem cannot be solved in a closed-form way. 
Thus, we propose to use a sampling-based optimization method \cite{homem2014monte} that consists in sampling different informative state trajectories $\stateInfTraj$ and choose the one that shows the smallest informative cost.
However, this approach cannot be directly employed due to some practical issues:
\begin{enumerate}
	\item To compute the informative cost for every sampled informative trajectory we should theoretically run an experiment. This is clearly time consuming and does not meet the goals of Objective~\ref{obj}.
	\item We do not know $\inputTaskTraj$. From its definition, we should know $\trueNonlinearDyn$ to compute $\inputTaskTraj$ given $\stateTaskTraj$. Therefore, we cannot directly compute $\learnedModelInputTaskDeltaSet$.
	\item It is not straightforward how we can compute the integral of the posterior variance over $\learnedModelInputTaskDeltaSet$.
	\item It is not straightforward how we can efficiently sample informative trajectories.
\end{enumerate}
Each of these four problems are individually addressed below  proposing a few approximations that make \eqref{eq:optimization:variance:ballTrackingTraj} solvable from a practical point of view.
This allows for deploying the method on real robots.

\subsection{Approximations of the optimization problem} 

\subsubsection{Approximation of the dynamical constraints}
\rev{During the search for the optimal informative state trajectory,
given a candidate informative state trajectory $\candidateStateInfTraj$,} instead of computing the posteriori variance based on the data collected from a real experiment, $\stateInputInformativeActualTraj$, we compute it based on the data collected from a simulation of the system, $\stateInputInfSimTraj$. %
In details, $\stateInputInfSimTraj$ is the output of the simulated closed-loop system using \rev{$\candidateStateInfTraj$} as reference trajectory, i.e.,
\begin{align}
	\begin{split}
		\stateInfSim[k+1] &= \modeledNonlinearDyn(\stateInfSim[k], \inputInfSim[k]) + \sampledResidualDynamics(\stateInputInfSimSample[k]) \\
  		\inputInfSim[k] &= \feedbackCtrl{ \rev{\candidateStateInf[k]}, \stateInfSim[k], \estimateNonlinearDyn},
  	\end{split}\label{eq:simulation_dynamics}
\end{align}
where $\sampledResidualDynamics(\stateInputInfSimSample[k])$ is a sample of the Bayesian model of the residual dynamics, using the probability distribution $\pdfs{\stateInputInfSimSample[k]}{\cdot, \cdot}$.
\rev{The bar $\bar{*}$ is used to denote quantities related to the simulation throughout this paper.}

\subsubsection{Approximation of $\learnedModelInputTaskDeltaSet$}\label{sec:approximations:learnedModelInputTaskDeltaSet}

Since we do not know $\trueNonlinearDyn$, we cannot compute $\inputTaskTraj$, and therefore neither $\learnedModelInputTaskDeltaSet$. 
In this section we show how we can get an estimation of $\learnedModelInputTaskDeltaSet$, denoted by $\learnedModelInputTaskDeltaSetEstimate$, exploiting the current estimation of $\trueNonlinearDyn$.

We firstly uniformly sample the state and input spaces, $\stateSet$ and $\inputSet$, creating the sets $\stateSet' \subset \stateSet$ and $\inputSet' \subset \inputSet$, respectively.
We then simulate the closed-loop system $M$ times using $\stateTaskTraj$ as reference, 
We obtain $M$ state and input trajectories $\stateInputTaskSimTrajSampled$ where 
\begin{align}
	\begin{split}
		\stateTaskSim\sampleIndex[k+1] &= \modeledNonlinearDyn(\stateTaskSim\sampleIndex[k], \inputTaskSim\sampleIndex[k]) + \sampledResidualDynamics(\stateInputTaskSimSample) \\
  		\inputTaskSim\sampleIndex[k] &= \feedbackCtrl{\stateTask[k], \stateTaskSim\sampleIndex[k], \estimateNonlinearDyn}.
  	\end{split}\label{eq:simulation_dynamics}
\end{align}

Finally, we compute $\learnedModelInputTaskDeltaSetEstimate$ as
\begin{align}
	\begin{split}
	\learnedModelInputTaskDeltaSetEstimate = \{ \stateInputPairInstance \in \stateSetSampled \times \inputSetSampled \, | \, \exists \, k\in (0,\ldots,N) \text{ and}\\
		 j\in(1,\ldots,M) \text{ s.t. } \norm{\stateInputPairInstance - \stateInputTaskSimSample[k]} \leq \epsilon\}.
	\end{split} 
	\label{eqn:learnedModelInputTaskDeltaSetEstimate}
\end{align}
Similar to \eqref{eq:optimization:BallDef}, the threshold $\epsilon$ is a heuristic that controls the exploration of the informative trajectory. With large $\epsilon$, the optimal trajectory should show a more exploratory behavior.
%

\subsubsection{Approximation of the informative cost}

We replace $\learnedModelInputTaskDeltaSet$ with $\learnedModelInputTaskDeltaSetEstimate$ in \eqref{eq:optimization:variance:ballTrackingTraj} and this integral can be approximately solved using numerical integration such as Monte-Carlo integration: we uniformly sample $S$ pairs $\sampledStateInputPair$ in $\learnedModelInputTaskDeltaSet$, where $j =1,\ldots,S$, creating the set $\learnedModelInputTaskDeltaSetSampled$.
We then approximate the informative cost in \eqref{eq:optimization:variance:ballTrackingTraj} as
\begin{equation} \label{eq:approximateIntegral}
 \frac{V}{S} \sum_{\sampledStateInputPair \in \learnedModelInputTaskDeltaSetEstimate'} \priorVariance{\sampledStateInputPair | \stateInputInformativeActualTraj},
\end{equation}
where $V$ is the volume $\int_{ \learnedModelInputTaskDeltaSetEstimate }d \stateInputPairInstance$.
Notice that for the different sampled informative trajectories, $V$ and $S$ remain constant and therefore can be omitted in the optimization.

\subsubsection{Parametrization of the informative trajectory}

Since we want to improve the knowledge of the model in $\learnedModelInputTaskDeltaSet$, it is natural to think that the informative state trajectory $\stateInfTraj$ should be ``close'' to the task state trajectory $\stateTaskTraj$.
Therefore, given a generic $k$, we define $\stateInf[k]$ such that
\begin{align}
	\begin{split}
	\stateInf[k] = \stateTask[k] + \dState[k]
	\end{split}.
	\label{eq:infTraj:parametrization}
\end{align}
Now, sampling informative state trajectories means sampling ``deviations'' from the task state trajectory.
To reduce the sampling space, which has the same dimension of $\stateSet$, we parametrize $\dState$ using the Discrete Fourier Transform (DFT)
\begin{align}
    \dState[k] = \frac{1}{P} \sum_{p=0}^{P-1} \stateInfParam^\top\ve{p} e^{j\frac{2\pi p}{P} k} \label{eq:infTrajDeviation},
\end{align}
where $P\in \N_{>0}$, $j$ is the complex operator, $\ve{p} \in \R^{P}$ is a vector with 1 in place $p$ and $0$ elsewhere, and $\stateInfParam \in \stateInfParamSet \subset \R^{n\times P}$ is the state  parameter matrix.

From sampling every state of the informative trajectory, we now samples only fewer parameters.
Furthermore, the rationale behind the use of DFT parametrization is that it gives us a more intuitive control of the frequencies of excitation. 
We can use fewer parameters to generate excitation signals that are spread through frequencies of interest. 
Intuitively, the deviation signal can be seen as an excitation signal added around the task state trajectory. 
As a result, the algorithm inherently explores locally around the task trajectory.

\subsection{Sampling-based optimization algorithm}\label{sec:optimizationAlgorithm}

Considering the previous simplifications, \eqref{eq:optimization:variance:ballTrackingTraj} becomes
\begin{equation}
	\begin{aligned}
		\min_{\stateInfParam} \quad & \sum_{\sampledStateInputPair \in \learnedModelInputTaskDeltaSetEstimate'} \priorVariance{\sampledStateInputPair | \stateInputInfSimTraj}	 \\
		\textrm{s.t.} \quad	
			& \learnedModelInputTaskDeltaSetEstimate \text{ as in \eqref{eqn:learnedModelInputTaskDeltaSetEstimate}}, \, \, \sampledStateInputPair \text{ as in \eqref{eq:simulation_dynamics}}, \\
			& \stateInf[k] = \stateTask[k] + \dState[k], \, \, \dState[k] \text{ as in \eqref{eq:infTrajDeviation}}.
	\end{aligned}
	\label{eq:optimization:variance:approximation}
\end{equation}

Practically, to solve \eqref{eq:optimization:variance:approximation} we used a Monte-Carlo sampling-based method.
The algorithm follows the next steps which require the simulation of the system only:
\begin{enumerate}
	\item\label{alg:sampleParam} Uniformly sample a set of parameters $\stateInfParam \in \stateInfParamSet$ and compute several informative state trajectories as in \eqref{eq:infTraj:parametrization} and \eqref{eq:infTrajDeviation};
	\item Simulate multiple times the system with the sampled residual model $\sampledResidualDynamics$ according to the prior model. Each informative state trajectories computed at step 1 is used as reference; 
	\item For every simulation, collect the data relative to the performed trajectory, $\stateInputInfSimTraj$, and update the Bayesian model of the residual dynamics;
	\item Compute $\learnedModelInputTaskDeltaSetEstimate$ as explained in \ref{sec:approximations:learnedModelInputTaskDeltaSet};
	\item Evaluate the information cost in $\learnedModelInputTaskDeltaSetEstimate$ according to \eqref{eq:approximateIntegral} associated to every new updated model;
	\item Select the informative state trajectory corresponding to the minimum information cost.
\end{enumerate}
Once the informative state trajectory supposed to provide the best model update is selected, it is used as reference in a real experiment. 
The relative collected data, $\stateInputInformativeActualTraj$, is then employed to update the prior model.
The full process can be repeated from step \ref{alg:sampleParam}), to find a new state informative trajectory that would allow to further improve the model accuracy, and in turn to reduce the tracking error.

\textit{Remark}: Note that the quality of the approximate solution depends on the quality of the prior model. Therein lies the purpose of this algorithm: Within each iteration, the quality of the prior model improves, and the solution to the approximated problem converges towards the true optimum. Consequently, this helps improving the prior model.

\section{Application to an aerial robot: the omav}\label{sec:applicationOMAV}
This section shows how the above framework is applied on an omnidirectional flying vehicle, called \textit{omav}~\cite{bodie2019omnidirectional}. 
%
The omav (\figref{pic:omav}) is an overactuated omnidirectional flying vehicle with six tiltable arms in a hexagonal arrangement. A coaxial rotor configuration is rigidly attached to the end of each arm. 
The rotation of each arm can be actively controlled by a servo motor, which results in a total of 18 actuators. 
Although the setup enhances the motion and interaction capabilities, aerodynamic disturbances among the rotors, unknown servo dynamics, backlashes, and other mechanical inaccuracy are difficult to be modeled and included in standard model-based controller. 
This makes omav a suitable testbed to validate the proposed method for active model learning.


The state of the omav is given by $\stateInstance = [\posState^\top \,\; \attState^\top \,\; \velState^\top \,\;  \angVelState^\top]^\top \in \stateSet \subset \R^{12}$. 
In order, $\stateInstance$ includes the position, attitude (expressed in Euler angles), linear velocity, and angular velocity of the vehicle.
As input of the system we consider the commanded wrench, i.e., the total force and moment commanded to the vehicle\footnote{For simplicity, we consider force and moment scaled by mass and inertia, respectively.}, $\inputInstance = [\forceVecCmd^\top \,\; \torqueVecCmd^\top]^\top \in \inputSet \subset \R^{6}$.
We assume that an allocation policy is implemented to transform $\inputInstance$ into low level commands for the servos and the motors \cite{bodie2019omnidirectional}.
Finally, the dynamics of the omav can be written as in \eqref{eq:dynamcis:true:decomposition}, where $\modeledNonlinearDyn$ is derived using standard Newton-Euler equations.
Notice that $\modeledNonlinearDyn$ is linear with respect to the input and can be written as
\begin{equation}
	\modeledNonlinearDyn(\stateInputPairInstance)= l(\stateInstance) + \inputInstance,
\end{equation}
 where  $l(\stateInstance)$ includes all the terms that do not depend on $\inputInstance$.
 
On the other hand, $\unkownNonlinearDyn$ includes all previously mentioned unmodeled dynamic behaviors that cannot be easily captured with first principles.
Considering the last six row of the dynamics (the linear and angular accelerations), we can consider $\unkownNonlinearDyn(\stateInputPairInstance)$ as the mismatch between the commanded wrench and the actuated one.

The controller tries to implement a feedback linearization control law with a PID action on the position and attitude errors.
In particular, given a reference task trajectory, $\stateTaskTraj$, and a priori model for $\unkownNonlinearDyn$, the controller $\feedbackCtrl{\stateTask[k], \discreteState{k}, \estimateNonlinearDyn}$ tries to find the input $\inputInstance[k]$ that solves the following optimization problem
\begin{equation}
	\min_{\inputInstance[k]} \norm{\stateInstance^\star[k] - l(\stateInstance[k]) - \inputInstance[k] - \estimateUnkownNonlinearDyn(\inputInstance[k])},
\end{equation}
where $\stateInstance^\star[k] = \mvec{K}(\stateTask[k] - \stateInstance[k]) + \mvec{K}_I \sum_{j=0}^k (\stateTask[j] - \stateInstance[j])$ is the PID action, with $\mvec{K}, \mvec{K}_I \in \R^{12 \times 12}$ positive definite matrices.
For the details about the implementation of such an optimization, we refer the interested reader to \cite{zhang2020learning}.

From experimental observations, we remark that the residual dynamics regarding the differential kinematics and linear acceleration (first nine rows) is almost negligible with respect to the one regarding the angular acceleration (last three rows).
In other words, the mismatch between commanded and actual force is much smaller than the one between commanded and actual torque. 
For this reason, in this first work, we focus on the attitude dynamics, applying the proposed active dynamics learning only on the last three rows of the system dynamics. These mismatches are modeled as three independent single-output Gaussian processes with $\inputInstance$ as the training input and the torque model mismatch as training output. We neglect the rotational drag torque acting on the vehicle \rev{since the vehicle is mostly operating with low angular velocities}, thus $\estimateUnkownNonlinearDyn$ is modeled independent of the state.


\section{Experimental results}


The experimental platform is the omnidirectional aerial vehicle in Fig.~\ref{pic:omav}: the omav. 
The omav weighs $4 \,\rm kg$ and is equipped with a NUC i7 computer and a PixHawk flight controller. 
This configuration allows to run all the necessary algorithms onboard implemented in a ROS framework.
A motion capture system provides pose estimates at $100\,\rm Hz$.
For a more complete description of the testbed see \cite{bodie2019omnidirectional}.

As stated in Section~\ref{sec:applicationOMAV}, the proposed method has been implemented and evaluated focusing on the rotational dynamics.
For the learned Gaussian process model, data points are subsampled from the experimental data using the $k$-medoids \cite{macqueen1967some} algorithm where the Euclidean squared distance between the inputs is used as the distance metric. Throughout the experiments, squared exponential kernels are used.
The deviation $\dState[k]$ is sampled around $x, y, z$-axis on the angular acceleration level, constraining to be below 2 Hz. Note that this is equivalent to giving $\dState[k]$ on the angular velocity. 
For simplicity, we limit the number of frequencies $P$ to 2 and allow the frequency locations to be sampled along with its magnitude. This yields a total of 12 coefficients to be sampled. 
The simulation framework is set up using RotorS Gazebo simulator \cite{Furrer2016}.
In this section we use ``non-informative trajectory'' to describe the case where only the task trajectory is used to collect the data to update the model.

\subsection{Correlation between informative cost and tracking error}
  
\begin{figure}[t]
\centering
\begin{minipage}[t]{.52\columnwidth}
\includegraphics[width=\columnwidth]{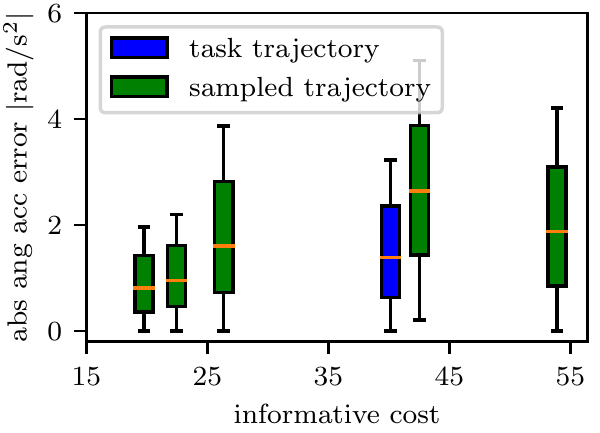}
\caption{A comparison of the tracking performance using the model learned from sampled trajectories and task trajectory. }
\label{pic:trackingErrorvsInformativeCost}
\end{minipage}\hfill
\begin{minipage}[t]{.46\columnwidth}
\includegraphics[width=\columnwidth]{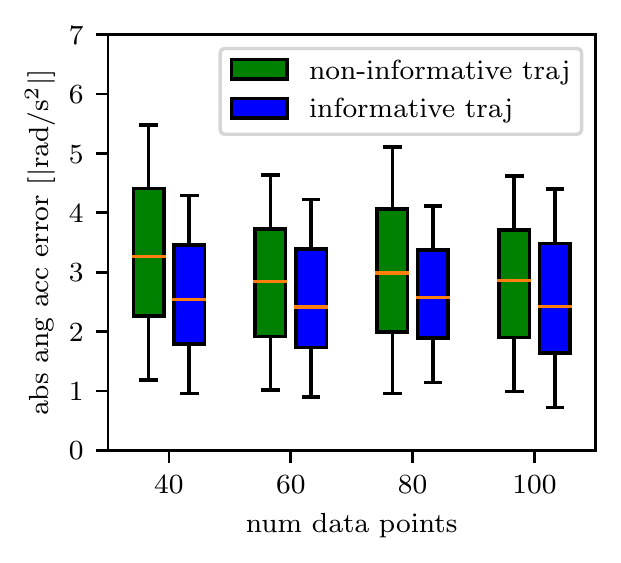}
\caption{A comparison of the tracking performance between informative trajectory and task trajectory for the same number of data points. 
  }
\label{pic:experiment_efficiency}
\end{minipage}
\end{figure}

  
An experiment is conducted to investigate whether the tracking error defined in problem \eqref{eq:optimization:trackingError} is correlated to the informative cost defined in \eqref{eq:optimization:variance:approximation}. 
The omav is asked to follow a pitching trajectory up to 60 degrees in pitch and 1 $\mathrm{rad}/s^2$ in pitch angular acceleration, similar to previous work \cite{zhang2020learning}. 
A prior model is built by collecting the data from executing the task trajectory.
Next, five sampled candidate informative trajectories and the task trajectory are executed and six learned models are built accordingly. 
They are then evaluated on the test data generated by the prior model. 
$\epsilon$ is heuristically tuned by simulation  computing the average distance between the commanded wrench and the achieved wrench.
The tracking performance of the angular acceleration\footnote{Notice that evaluating the angular acceleration tracking is equivalent to evaluate the error between actual and commanded torque which strongly depends on the model accuracy.} along the $y$-axis using these models are shown in \figref{pic:trackingErrorvsInformativeCost}. 
It can be seen that there is a clear correspondence between the informative cost and the tracking error. 
Furthermore, the model learned from the task trajectory does not yield the best tracking performance.
\subsection{Comparison between informative and task trajectory}
\begin{figure}[t]
\begin{minipage}[b]{.52\columnwidth}
  \centering
  \includegraphics[trim=10em 0 10em 0, clip,width=\columnwidth]{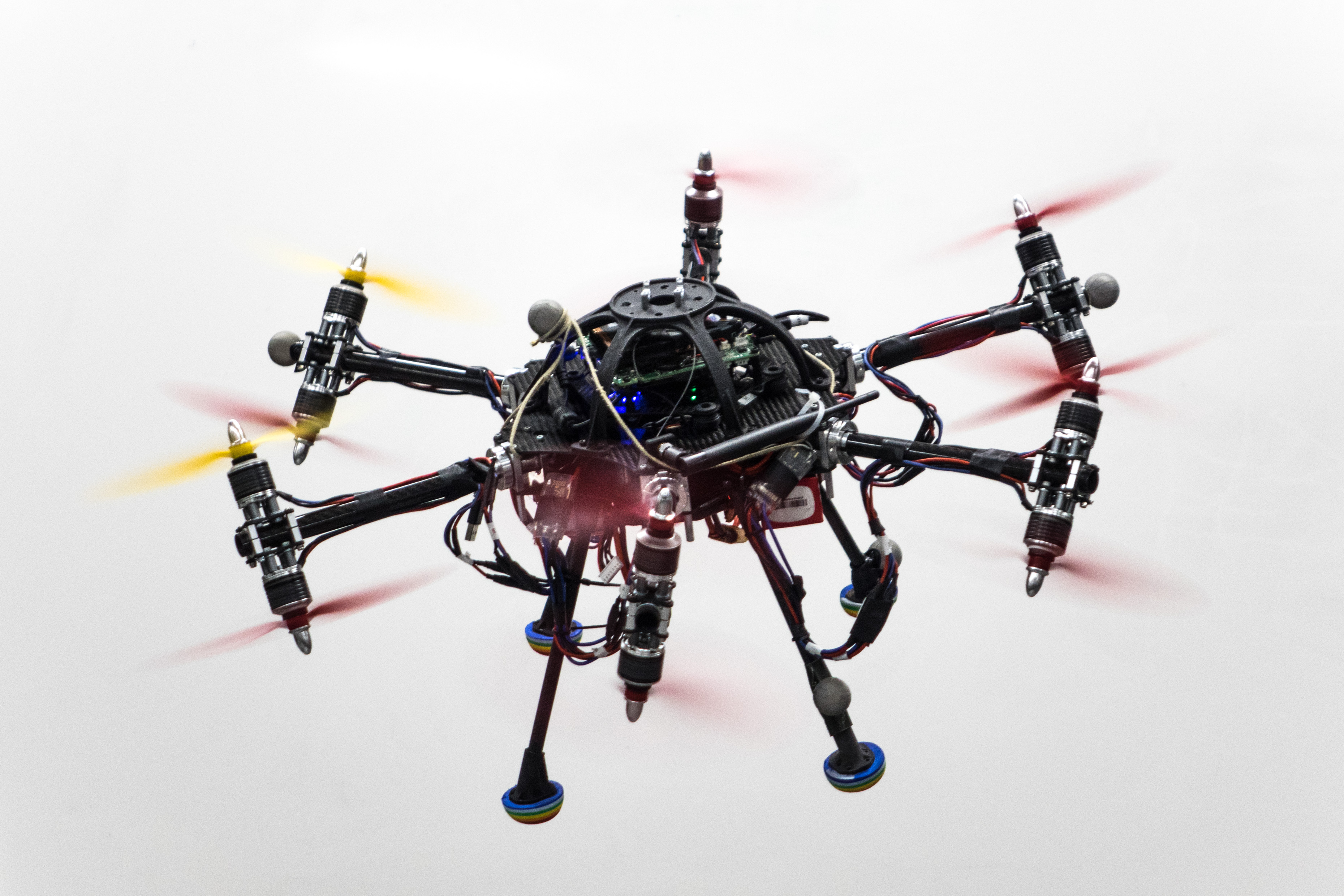}
  \caption{The omnidirectional flying vehicle (omav) used to experimentally validate our method.}
  \label{pic:omav}
\end{minipage}
 \hfill
%
\begin{minipage}[b]{.46\columnwidth}
\centering
\includegraphics[width=\columnwidth]{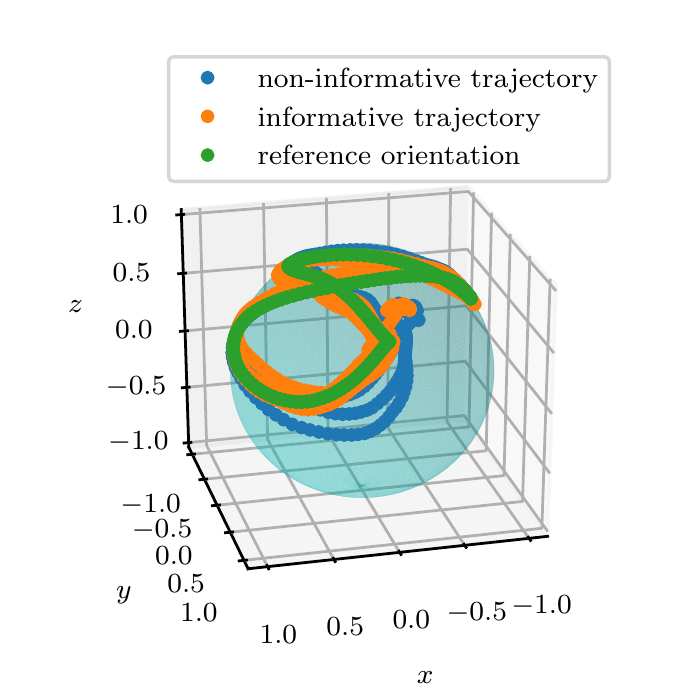}
\caption{Tracking of a body-fixed unit vector $(1,1,1)/\sqrt{3}$ is plotted on a unit sphere.}
\label{pic:so3Tracking}
\end{minipage}



\end{figure}

To compare the efficiency of the informative and non-informative trajectory,
a figure-8 in attitude (with roll and pitch up to 26 degrees) with constant position is given as a task trajectory (see \figref{pic:so3Tracking}). 
We compute the prior model running the task trajectory for the first time.
Then 20 trajectories are randomly generated and evaluated in simulation as explained in Section~\ref{sec:optimizationAlgorithm} using the prior model.
The most informative trajectory (lowest informative cost) and the task trajectory are then executed and the data are recorded for both trajectories. 
We subsampled 20, 40, 60, 80 data points from the experiments running each trajectory and built a model for each of these combinations by augmenting the prior model with these data points. The hyperparameters of the Gaussian processes are reoptimized. 
The models are then used in the controller to track the task trajectory in real experiments for validation. 
Tracking performance are evaluated in \figref{pic:experiment_efficiency} as the average of the absolute angular acceleration over all three axes.
It can be noted that for the same amount of data points, the informative trajectory always outperforms the non-informative trajectory in term of both mean tracking error and corresponding variance.
On average the performance\footnote{By performance of a trajectory we mean the tracking performance using the updated controller with the data collected from that trajectory.} of informative trajectories outperforms the non-informative one by 13.3\%.  


\begin{figure}[t]
\centering
\includegraphics[width=\columnwidth]{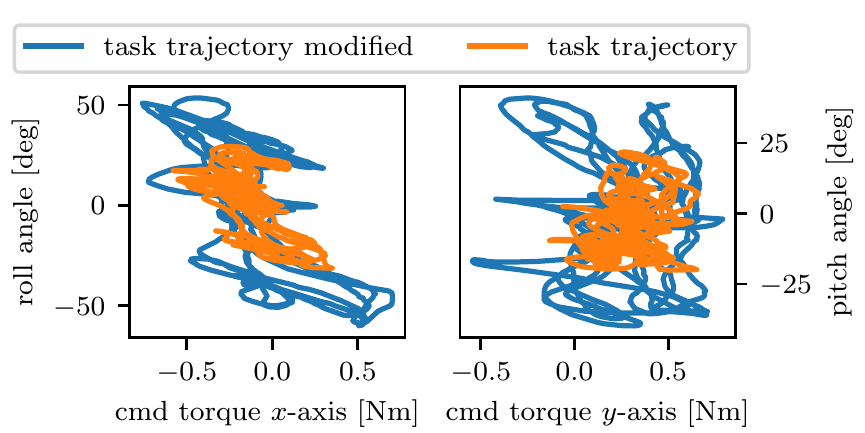}
\caption{Phase plots of the task trajectory and modified task trajectory. It can be observed that although the modified trajectory extend beyond the task trajectory, the model learned from the informative trajectory helps to reduce the tracking.}
\label{pic:phasePlot}
\end{figure}

\subsection{Comparison of the model generalizability}
To test the generalizability of the model learned from the informative trajectory, a modified figure-8 trajectory with higher pitch and roll reference angles (up to 43 degrees) is used. 
As can be seen in the phase plot in \figref{pic:phasePlot}, the state input pairs of the modified figure-8 extend up to twice of the original one.
In this case, both models from the informative trajectory and the non-informative trajectory have 100 data points. 
It can be seen from Table \ref{tab:generalizability} that the model learned from informative trajectory yields better tracking performance, especially around the $z$-axis.
\begin{table}[t!]
\renewcommand{\arraystretch}{1.3}
\centering
\begin{tabular}{c|c|c|c}
\hline
 & $x$-axis &  $y$-axis  &  $z$-axis   \\
\hline
non-informative&  38.4\%  & 41.7\%  & 23\% \\
\hline
informative &  43.2\% &  57.9 \% & 62\% \\
\hline
\end{tabular}
\caption{Angular acceleration tracking error reduction with respect to the case without model learning in percentage.}
\label{tab:generalizability}
\end{table}

\section{Conclusion}
This work presents a practical framework that effectively and efficiently collects data points for the learning of models used at the control level to significantly improve tracking performance on real robots.
We experimentally demonstrate the validity of the method on an overactuated aerial robot, the omav, whose dynamics is complex  and difficult to learn.
Experimental results show that the learned model from informative trajectories is efficient in data points collection and generalizes on modified trajectories.



%
\IEEEpeerreviewmaketitle

\bibliographystyle{IEEEtran}
\bibliography{IEEEabrv,bibliography}

\begin{thebibliography}{10}
\providecommand{\url}[1]{#1}
\csname url@samestyle\endcsname
\providecommand{\newblock}{\relax}
\providecommand{\bibinfo}[2]{#2}
\providecommand{\BIBentrySTDinterwordspacing}{\spaceskip=0pt\relax}
\providecommand{\BIBentryALTinterwordstretchfactor}{4}
\providecommand{\BIBentryALTinterwordspacing}{\spaceskip=\fontdimen2\font plus
\BIBentryALTinterwordstretchfactor\fontdimen3\font minus
  \fontdimen4\font\relax}
\providecommand{\BIBforeignlanguage}[2]{{%
\expandafter\ifx\csname l@#1\endcsname\relax
\typeout{** WARNING: IEEEtran.bst: No hyphenation pattern has been}%
\typeout{** loaded for the language `#1'. Using the pattern for}%
\typeout{** the default language instead.}%
\else
\language=\csname l@#1\endcsname
\fi
#2}}
\providecommand{\BIBdecl}{\relax}
\BIBdecl

\bibitem{abbeel2010autonomous}
P.~Abbeel, A.~Coates, and A.~Y. Ng, ``Autonomous helicopter aerobatics through
  apprenticeship learning,'' \emph{The International Journal of Robotics
  Research}, vol.~29, no.~13, pp. 1608--1639, 2010.

\bibitem{kamel2017model}
M.~Kamel, T.~Stastny, K.~Alexis, and R.~Siegwart, ``Model predictive control
  for trajectory tracking of unmanned aerial vehicles using robot operating
  system,'' in \emph{Robot operating system (ROS)}.\hskip 1em plus 0.5em minus
  0.4em\relax Springer, 2017, pp. 3--39.

\bibitem{kuindersma2016optimization}
S.~Kuindersma, R.~Deits, M.~Fallon, A.~Valenzuela, H.~Dai, F.~Permenter,
  T.~Koolen, P.~Marion, and R.~Tedrake, ``Optimization-based locomotion
  planning, estimation, and control design for the atlas humanoid robot,''
  \emph{Autonomous robots}, vol.~40, no.~3, pp. 429--455, 2016.

\bibitem{ostafew2014learning}
C.~J. Ostafew, A.~P. Schoellig, and T.~D. Barfoot, ``Learning-based nonlinear
  model predictive control to improve vision-based mobile robot path-tracking
  in challenging outdoor environments,'' in \emph{IEEE International Conference
  on Robotics and Automation}, 2014.

\bibitem{gillespie2018learning}
M.~T. Gillespie, C.~M. Best, E.~C. Townsend, D.~Wingate, and M.~D. Killpack,
  ``Learning nonlinear dynamic models of soft robots for model predictive
  control with neural networks,'' in \emph{2018 IEEE International Conference
  on Soft Robotics (RoboSoft)}.\hskip 1em plus 0.5em minus 0.4em\relax IEEE,
  2018, pp. 39--45.

\bibitem{bodie2019omnidirectional}
K.~Bodie, M.~Brunner, M.~Pantic, S.~Walser, P.~Pfndler, U.~Angst, R.~Siegwart,
  and J.~Nieto, ``{An Omnidirectional Aerial Manipulation Platform for
  Contact-Based Inspection},'' \emph{Robotics: Science and Systems XV}, 2019.

\bibitem{kabzan2019learning}
J.~Kabzan, L.~Hewing, A.~Liniger, and M.~N. Zeilinger, ``Learning-based model
  predictive control for autonomous racing,'' \emph{IEEE Robotics and
  Automation Letters}, 2019.

\bibitem{zhang2020learning}
W.~Zhang, M.~Brunner, L.~Ott, M.~Kamel, R.~Siegwart, and J.~Nieto, ``Learning
  dynamics for improving control of overactuated flying systems,'' \emph{IEEE
  Robotics and Automation Letters}, vol.~5, no.~4, pp. 5283--5290, 2020.

\bibitem{nguyen2010using}
D.~Nguyen-Tuong and J.~Peters, ``Using model knowledge for learning inverse
  dynamics,'' in \emph{IEEE International Conference on Robotics and
  Automation}, 2010.

\bibitem{hwangbo2019learning}
J.~Hwangbo, J.~Lee, A.~Dosovitskiy, D.~Bellicoso, V.~Tsounis, V.~Koltun, and
  M.~Hutter, ``Learning agile and dynamic motor skills for legged robots,''
  \emph{Science Robotics}, 2019.

\bibitem{ljung1999system}
L.~Ljung, \emph{System identification, Theory for the user}.\hskip 1em plus
  0.5em minus 0.4em\relax Prentice Hall, 1999.

\bibitem{settles2009active}
B.~Settles, ``Active learning literature survey,'' University of
  Wisconsin-Madison Department of Computer Sciences, Tech. Rep., 2009.

\bibitem{schrangl2020iterative}
P.~Schrangl, P.~Tkachenko, and L.~del Re, ``Iterative model identification of
  nonlinear systems of unknown structure: Systematic data-based modeling
  utilizing design of experiments,'' \emph{IEEE Control Systems Magazine},
  vol.~40, no.~3, pp. 26--48, 2020.

\bibitem{wilson2016dynamic}
A.~D. Wilson, J.~A. Schultz, A.~R. Ansari, and T.~D. Murphey, ``Dynamic task
  execution using active parameter identification with the baxter research
  robot,'' \emph{IEEE Transactions on Automation Science and Engineering},
  vol.~14, no.~1, pp. 391--397, 2016.

\bibitem{koller2018learning}
T.~Koller, F.~Berkenkamp, M.~Turchetta, and A.~Krause, ``Learning-based model
  predictive control for safe exploration,'' in \emph{2018 IEEE Conference on
  Decision and Control (CDC)}.\hskip 1em plus 0.5em minus 0.4em\relax IEEE,
  2018, pp. 6059--6066.

\bibitem{buisson2019actively}
M.~Buisson-Fenet, F.~Solowjow, and S.~Trimpe, ``Actively learning gaussian
  process dynamics,'' \emph{arXiv preprint arXiv:1911.09946}, 2019.

\bibitem{zimmer2018safe}
C.~Zimmer, M.~Meister, and D.~Nguyen-Tuong, ``Safe active learning for
  time-series modeling with gaussian processes,'' in \emph{Advances in Neural
  Information Processing Systems}, 2018, pp. 2730--2739.

\bibitem{nakka2020chance}
Y.~K. Nakka, A.~Liu, G.~Shi, A.~Anandkumar, Y.~Yue, and S.-J. Chung,
  ``Chance-constrained trajectory optimization for safe exploration and
  learning of nonlinear systems,'' \emph{arXiv preprint arXiv:2005.04374},
  2020.

\bibitem{borrelli2017predictive}
F.~Borrelli, A.~Bemporad, and M.~Morari, \emph{Predictive control for linear
  and hybrid systems}.\hskip 1em plus 0.5em minus 0.4em\relax Cambridge
  University Press, 2017.

\bibitem{capone2020localized}
A.~Capone, G.~Noske, J.~Umlauft, T.~Beckers, A.~Lederer, and S.~Hirche,
  ``Localized active learning of gaussian process state space models,'' in
  \emph{Learning for Dynamics and Control}.\hskip 1em plus 0.5em minus
  0.4em\relax PMLR, 2020, pp. 490--499.

\bibitem{congdon2007bayesian}
P.~Congdon, \emph{Bayesian statistical modelling}.\hskip 1em plus 0.5em minus
  0.4em\relax John Wiley \& Sons, 2007, vol. 704.

\bibitem{homem2014monte}
T.~Homem-de Mello and G.~Bayraksan, ``Monte carlo sampling-based methods for
  stochastic optimization,'' \emph{Surveys in Operations Research and
  Management Science}, vol.~19, no.~1, pp. 56--85, 2014.

\bibitem{macqueen1967some}
J.~MacQueen \emph{et~al.}, ``Some methods for classification and analysis of
  multivariate observations,'' in \emph{Proceedings of the fifth Berkeley
  symposium on mathematical statistics and probability}, vol.~1, no.~14.\hskip
  1em plus 0.5em minus 0.4em\relax Oakland, CA, USA, 1967, pp. 281--297.

\bibitem{Furrer2016}
\BIBentryALTinterwordspacing
F.~Furrer, M.~Burri, M.~Achtelik, and R.~Siegwart, \emph{Robot Operating System
  (ROS): The Complete Reference (Volume 1)}.\hskip 1em plus 0.5em minus
  0.4em\relax Cham: Springer International Publishing, 2016, ch. RotorS---A
  Modular Gazebo MAV Simulator Framework, pp. 595--625. [Online]. Available:
  \url{http://dx.doi.org/10.1007/978-3-319-26054-9_23}
\BIBentrySTDinterwordspacing

\end{thebibliography}
%

\end{document}